\documentclass[10pt, conference, compsocconf]{IEEEtran}

\usepackage{subfigure}
\usepackage{graphicx}
\usepackage{multirow}

\usepackage{amsmath,amssymb}
\usepackage{algorithm,algpseudocode}
\usepackage{booktabs,balance}
\usepackage{bm}
\usepackage{color}
\usepackage{courier}
\usepackage{epsfig}
\usepackage{graphicx}
\usepackage{helvet}
\usepackage{multirow}
\usepackage{times}
\usepackage{url}
\usepackage{cite}
\usepackage{lipsum}

\newcommand\ie{\emph{i.e.}}

\newcommand\eg{\emph{e.g.}}

\ifCLASSINFOpdf
\else
\fi

\begin{document}
\title{Cascaded Detail-Preserving Networks for Super-Resolution of Document Images}

\author{
\IEEEauthorblockN{
Zhichao Fu$^{1}$,
Yu Kong$^{2}$,
Yingbin Zheng$^{2}$,
Hao Ye$^{2}$,
Wenxin Hu$^{1}$,
Jing Yang$^{1}$,
Liang He$^{1}$}
\\
$^{1}$East China Normal University, Shanghai, China~~~~$^{2}$Videt Tech, Shanghai, China
}

\maketitle

\begin{abstract}

The accuracy of OCR is usually affected by the quality of the input document image and different kinds of marred document images hamper the OCR results. Among these scenarios, the low-resolution image is a common and challenging case. In this paper, we propose the cascaded networks for document image super-resolution. Our model is composed by the \emph{Detail-Preserving Networks} with small magnification. The loss function with perceptual terms is designed to simultaneously preserve the original patterns and enhance the edge of the characters. These networks are trained with the same architecture and different parameters and then assembled into a pipeline model with a larger magnification. The low-resolution images can upscale gradually by passing through each Detail-Preserving Network until the final high-resolution images. Through extensive experiments on two scanning document image datasets, we demonstrate that the proposed approach outperforms recent state-of-the-art image super-resolution methods, and combining it with standard OCR system lead to signification improvements on the recognition results.

\end{abstract}

\IEEEpeerreviewmaketitle

\newcommand\blfootnote[1]{
\begingroup
\renewcommand\thefootnote{}\footnote{#1}
\addtocounter{footnote}{-1}
\endgroup
}

\blfootnote{
This work was supported in part by the Science and Technology Commission of Shanghai Municipality under Grant 18511103105. Zhichao Fu and Yu Kong contributed equally to this work. Corresponding author: Jing Yang (e-mail: jyang@cs.ecnu.edu.cn).}

\section{Introduction}
\label{sec:intro}

Image super-resolution (SR) is an important and challenging low-level vision task in many real-world problems.
In this paper, we focus on the application of super-resolution for the document images, which are one of the most pervasive types of input in our daily life~\cite{mancas2007introduction}. The document images with low-quality can affect the results of OCR and lead to low OCR accuracy. There are different kinds of marred document inputs, and the low-resolution images are a common case among these scenarios. In order to improve the OCR accuracy, super-resolution is usually considered as a pre-processing enhancement stage.

Super-resolution involves adding details and keeping a smooth structure based on the original low-resolution images (LR). It is a typical ill-posed problem to predict those unseen pixels for the real high-resolution images (HR) \cite{arxiv:1902.06068}.
Traditional super-resolution methods usually employ the interpolation based approach such as Bilinear and Bicubic. Recently, the applications of deep learning and generative networks on computer vision research have created a significant breakthrough in many fields. For super-resolution of the natural images, the deep models such as SRCNN~\cite{Chao2014Learning,Dong2016Image} and SRGAN~\cite{Ledig2016Photo} have achieved state-of-the-art performance.
However, natural images and document images contain different attributes, and the reasons to have low-resolution images are also different.
The results of the previous method tend to improve the overall similarity with the HR images, which sometimes cause blurry edges and cannot bring improvement to the OCR accuracy.

\begin{figure}[t]
\centering
\includegraphics[width=.9\linewidth]{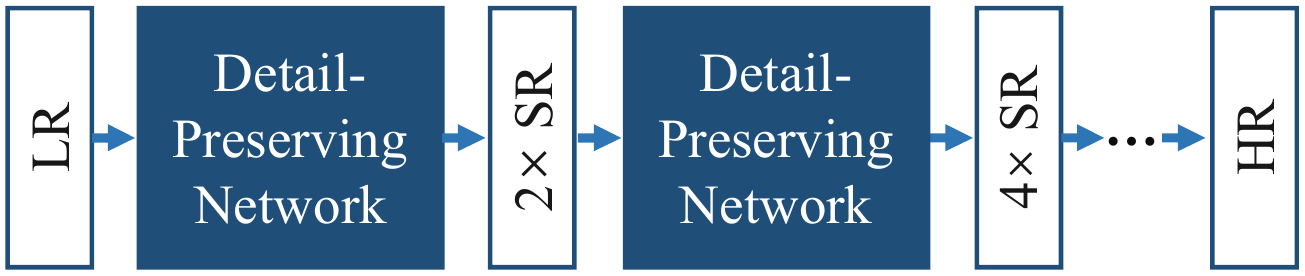}
\caption{Pipeline of proposed Cascaded Detail-Preserving Networks.}
\label{fig:framework}
\end{figure}

Many previous methods use a single network with continuous up-sample blocks after the convolution blocks. After one single up-sample process, the intermediate image features may not be adequately extracted and the text regions under low-resolution may be processed into unrecognizable characters for the OCR system.
In this paper, we propose to use the cascaded networks and the pipeline is illustrated in Fig. \ref{fig:framework}. Each \emph{Detail-Preserving Networks} (\emph{DPNet}) aims to preserve the detail with small magnification. They are trained with the same architecture and different parameters and then assembled into a pipeline model with a larger magnification. The low-resolution images can upscale gradually by passing through each DPNet until the final high-resolution images. For each DPNet, the loss function with perceptual terms is designed to simultaneously preserve the content and enhance the edge of the characters.
We conduct extensive experiments with state-of-the-art image super-resolution methods on two scanning document image datasets and demonstrate its superiority in terms of \emph{Peak Signal to Noise Ratio} (PSNR) and \emph{Structural Similarity Index Measure} (SSIM)~\cite{Wang2004SSIM} over previous approaches. Besides, combining our Cascaded Detail-Preserving Networks framework with standard OCR system also lead to signification improvements on the recognition results.

The rest of this paper is organized as follows. Section \ref{sec:related} introduces the background of super-resolution.
Section \ref{sec:Proposed_Model} discusses the model design, network architecture and training process in detail.
In Section \ref{sec:exp}, we demonstrate the qualitative and quantitative study of the proposed network. And we conclude our work in Section \ref{sec:conclusion}.

\begin{figure*}[t]
\centering
\includegraphics[width=.8\linewidth]{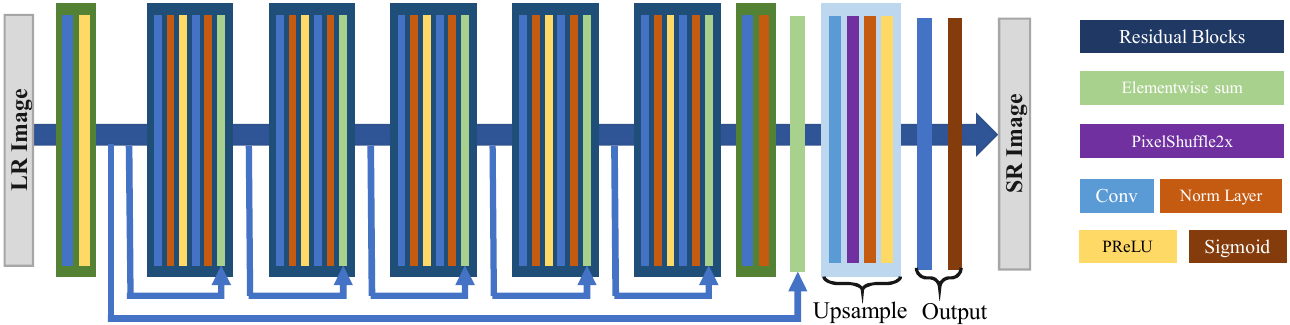}
\caption{Structure of Detail-Preserving Network.}
\label{fig:SRUnitNet architecture}
\end{figure*}

\section{Related Work}
\label{sec:related}
Super-resolution is a typical image restoration task, aiming to convert the low-resolution images into high-resolution.
Super-resolution can be useful for many applications, especially for \emph{optical character recognition (OCR)}. Specifically, the loss of image details can seriously affect both text detection and recognition from the document images. Therefore, the super-resolution methods are usually introduced as a pre-processing step and can lead to improvement of a modern OCR system.

Image super-resolution is an ill-posed problem and the super-resolution of document images is a domain-specific task. Traditional super-resolution approach can be addressed by using priors. These methods include prediction based approach~\cite{Tomer2014A}, gradient profile-based approach~\cite{Jian2008Image}, image statistics based approach~\cite{Efrat2013Accurate, Fernandez2013Super}, patch-based models~\cite{Qiang2005Patch, Aodha2012Patch}, and external learning or example-based super resolution~\cite{Freeman2002Example}.

In recent years advances in deep learning benefit the vision problems. A set of models have been built for super-resolution using deep convolutional neural networks (CNN). For instance, \cite{Chao2014Learning} and \cite{Dong2016Image} proposed a CNN based method to super-resolve natural images, by using the network to learn the mapping between interpolated bicubic images from LR images and corresponding HR images. VDSR network~\cite{Kim2016Accurate} is designed to predict the residuals instead of pixel values with fast convergence speed. With a deeply-recursive convolutional network architecture, DRCN~\cite{Kim2015Deeply} reported a high performance with fewer model parameters. More recently, SRGAN~\cite{Ledig2016Photo} introduced residual network for single image super-resolution (SISR) and combined generative adversarial network (GAN). GAN based method extracts texture features from images by a deep CNN, such as VGG-16~\cite{simonyan2014deep}, and makes the super-resolved images have proper texture and good perceptual quality. The discriminator network also makes the super-resolution network learn the capacity for transferring low-resolution images into high-resolution images with details.

\section{Framework}
\label{sec:Proposed_Model}

The resolution of document images is an important factor for both OCR system and human vision to recognize text and characters. As a general rule, the lower the text resolution is, the more visual information lost, and the lower recognition accuracy will be reached. Besides, extremely high resolution may not bring higher accuracy but higher computation burden. Therefore, considering real-world OCR applications, the super-resolution model should have an adjustable magnification to handle varying degrees of low-resolution in text patches. If text resolution is especially low, the model should proceed with higher magnification. And as a preprocessing step,
an efficient super-resolution model is helpful for the whole OCR pipeline.
This motivates us to design light-weight network architecture and further build our composable model.

The goal of our framework is to super-resolve document images and text patches with adjustable magnification. It is designed to work as a cascade process. As shown in Fig. \ref{fig:framework}, the total model is composed of multiple networks. Each DPNet is with small super-resolution magnification (2$\times$). The networks trained for different scale of document images share the same network architecture but have different parameters. The whole model is connected with the DPNet trained from the neighboring scales. The input low-resolution image is magnified successively, results in a multiplicative magnified high-resolution image.

\subsection{Detail-Preserving Network}
\label{subsec:UnitSRNet Architecture}

As shown in Fig. \ref{fig:SRUnitNet architecture}, the Detail-Preserving Network employs a generative CNN architecture, which follows a common single image super-resolution pattern and includes three parts.
The first part is to extract features with constant size as the input image. Here we use a single convolutional layer with a kernel size of 9 to make low-level feature mapping from the input image. Then ${N}$ residual blocks will extract high-level features from a low-level feature map.
Here we choose ${N}=5$ in our experiments for a trade-off between the performance and the model efficiency, and a kernel size of 3 for the convolutional layers.  Skip connection is also included in this part and contributes to the residual blocks training and feature fusion between low-level and high-level.
The second part is the upsampling. Using a series of upsample blocks cannot make the most of feature between each scales, so we employ a single upsample block with sub-pixel convolutional layer\footnote{Suppose the magnification of the upsample block is 2, the single channel input size is $W\times H$, and input/output channel number is $CI$/$CO$. Given the input $CI\times W\times H$, the convolutional layer will generate a $4CO \times W\times H$ matrix, which then will be converted to an output of $CO\times 2W\times 2H$ by the pixel shuffle operation.}~\cite{Shi2016Real}. The final part is to generate the output map, including a single convolutional layer and sigmoid function.

Blocks in our network are chosen with the same type of normalization layer and activation layer.
Parametric ReLU~\cite{He2015Delving} is used in the activation layers, and batch normalization~\cite{ioffe2015batch} is used in the normalization layers.

\begin{figure}[t]
\centering
\includegraphics[width=.8\linewidth]{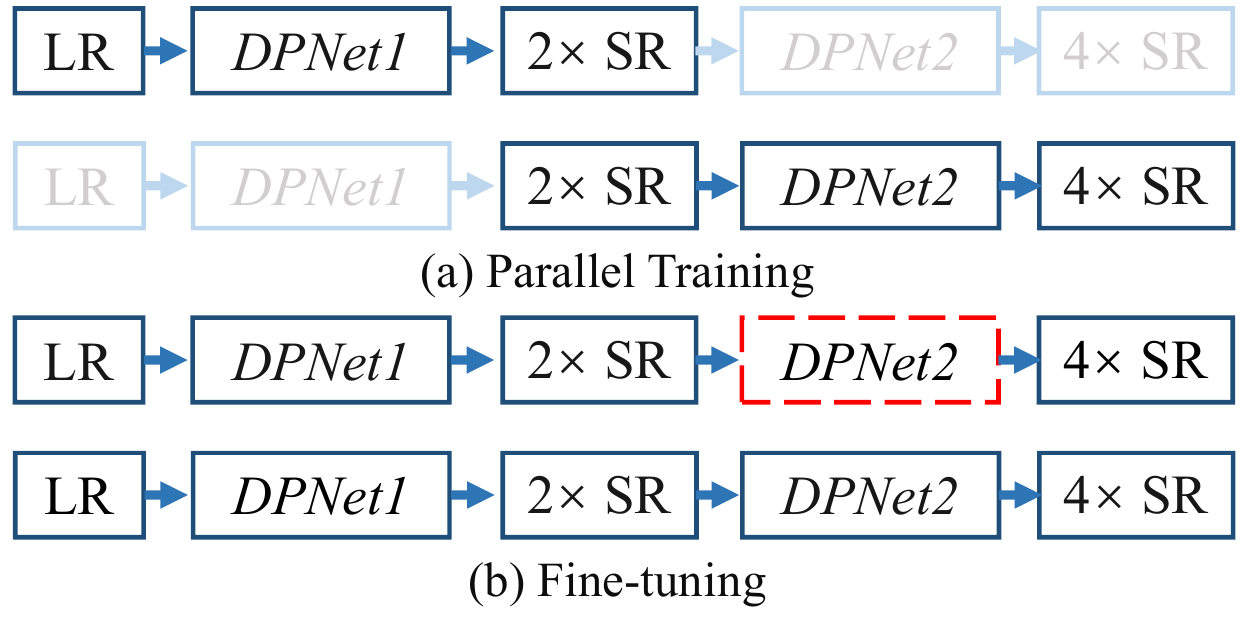}
\caption{Strategies to the training of the Cascaded DPNets. Red Block indicates the DPNet with frozen weights.}
\label{fig:training}
\end{figure}

\subsection{Model Training}
\label{sec:training}

Due to the cascade structure in this work, we divide the training process into two phases, \ie, parallel training and penetrating fine-tuning. An overview of the model training strategy is shown in Fig. \ref{fig:training}.

\subsubsection{Parallel Training}
\label{subsec:unit training}

Each network takes the image with lower resolution as the input and returns the images with higher resolution.
In the first phase, we suppose the networks for different scales are independent and trained them separately. Here we choose a 4$\times$ model as an example in Fig. \ref{fig:training}(a).
After down-sampling, 2$\times$ and 4$\times$ low-resolution images are generated from original high-resolution images.
4$\times$ low-resolution images are the input to \emph{DPNet1}. The outgoing super-resolved images are used to calculate loss with 2$\times$ low-resolution images. Then loss backward propagation will optimize parameters in this network.

In a similar way, \emph{DPNet2} is trained in parallel, using 2$\times$ low-resolution and original high-resolution images. Concerning the model with larger magnification, the networks can be trained paralleled in the same way, which are convenient when multiple GPUs are available.

\subsubsection{Fine-tuning}

The parallel training in the previous phase enables each DPNet to super-resolve images successfully with a small magnification. However, image restoration tasks are ill-posed problems and any model may not quickly find a perfect solution equal to the original high-resolution image. Therefore, we design this phase to adapt network parameters in Fig. \ref{fig:training}(b).

In each step, all of the networks connected by arrows are used for fine-tuning. The parameter weights of \emph{DPNet2} are initially frozen and the whole model takes low-resolution images as the input then outputs super-resolved 4$\times$ images to update the weight of \emph{DPNet1}.
The networks are fine-tuned sequentially in this phase, from the second to the \emph{N}-th (\eg, parameters of \emph{DPNet1} and \emph{DPNet2} are updated in Fig. \ref{fig:training}(b)).

\subsection{Loss Function}
\label{fig:loss}

For each phase and each network, the network employs the same loss function. Three terms are incorporated in the loss function as follows,
$$\mathcal{L}=\lambda_1\cdot\mathcal{L}_{Pixel}+\lambda_2\cdot\mathcal{L}_{Perceptual}+\lambda_3\cdot\mathcal{L}_{Edge}$$

The first term of the loss function is the pixel loss, which is defined by the pixel-wise MSE. Inspired by \cite{johnson2016perceptual}, The second term is the perceptual loss, which is based on the difference of feature maps from an ImageNet~\cite{ILSVRC15} pre-trained VGG19 network \cite{simonyan2014deep} between the generated and target image. Formally, the perceptual loss is defined as:
$$\mathcal{L}_{Perceptual}=
    \frac{1}{W_jH_j}
    \sum\limits_x^{W_j}
    \sum\limits_y^{H_j}
    (\phi_j(I_{HR})_{x,y}-\phi_j(S(I_{LR}))_{x,y})^2,
$$
where $I_{HR}$ and $I_{LR}$ indicate the high-resolution and low-resolution images, $\phi_j$ represents the $j$-th layer that outputs the feature maps with size ($W_j,H_j$), and $S(\cdot)$ is the super-resolution function. We choose the feature maps before the activation layer.
Both the pixel and perceptual terms represent the content of the images. Here we use the $L_2$ metric, as we found in our early experiments that the network trained using perceptual loss only or $L_1$ metric may cause unrealistic textures on generated images (which is also reported in previous work such as \cite{mechrez2018contextual}).

The last term is the edge loss. Here we employ the class-balanced cross-entropy loss~\cite{Xie2015Holistically}, by mapping the original high-resolution image and super-resolved image into the corresponding edge maps with holistically-nested edge detection (HED)~\cite{Xie2015Holistically}, and then computing their loss.
The benefits of the edge loss are two-folder. First, the enhancement of the edge information is able to preserve the detail information with small magnification. Second, as observed from the experiment, incorporating the edge loss accelerates the convergence speed during the model training.
The loss function is defined as
$$\mathcal{L}_{Edge}=
    l_{side}(\phi_{side_i}(I_{HR}),\phi_{side_i}(S(I_{LR}))),
$$
where $\phi_{side_i}$ is the edge maps from the $i$-th side-output layer of the network and $l_{side}$ indicates the class-balanced cross-entropy loss. We set $i=1$ in HED model to reduce the training and inference time. 

\subsection{Implementation Details}
\label{subsec:detail}

We implement our model using PyTorch\footnote{https://pytorch.org/}. The experiments are conducted using Intel Xeon-E5 CPU and NVIDIA Titan Xp GPUs. We evaluate some different methods with different fine-tuned network parameters but the same training dataset and configuration.

Adam solver\cite{Kingma2014Adam} is used for our model training on each network with parameters $\beta_1=0.9$ and $\beta_2=0.999$. The initial learning rate is 0.001 and decay to one-tenth every 20 epochs. As two-phase training is defined in Section \ref{sec:training}, we use 50 epochs for unit training and 5 epochs for fine-tuning training.

\section{Experiments}
\label{sec:exp}

\begin{figure*}[t]
\centering
\includegraphics[width=.8\linewidth]{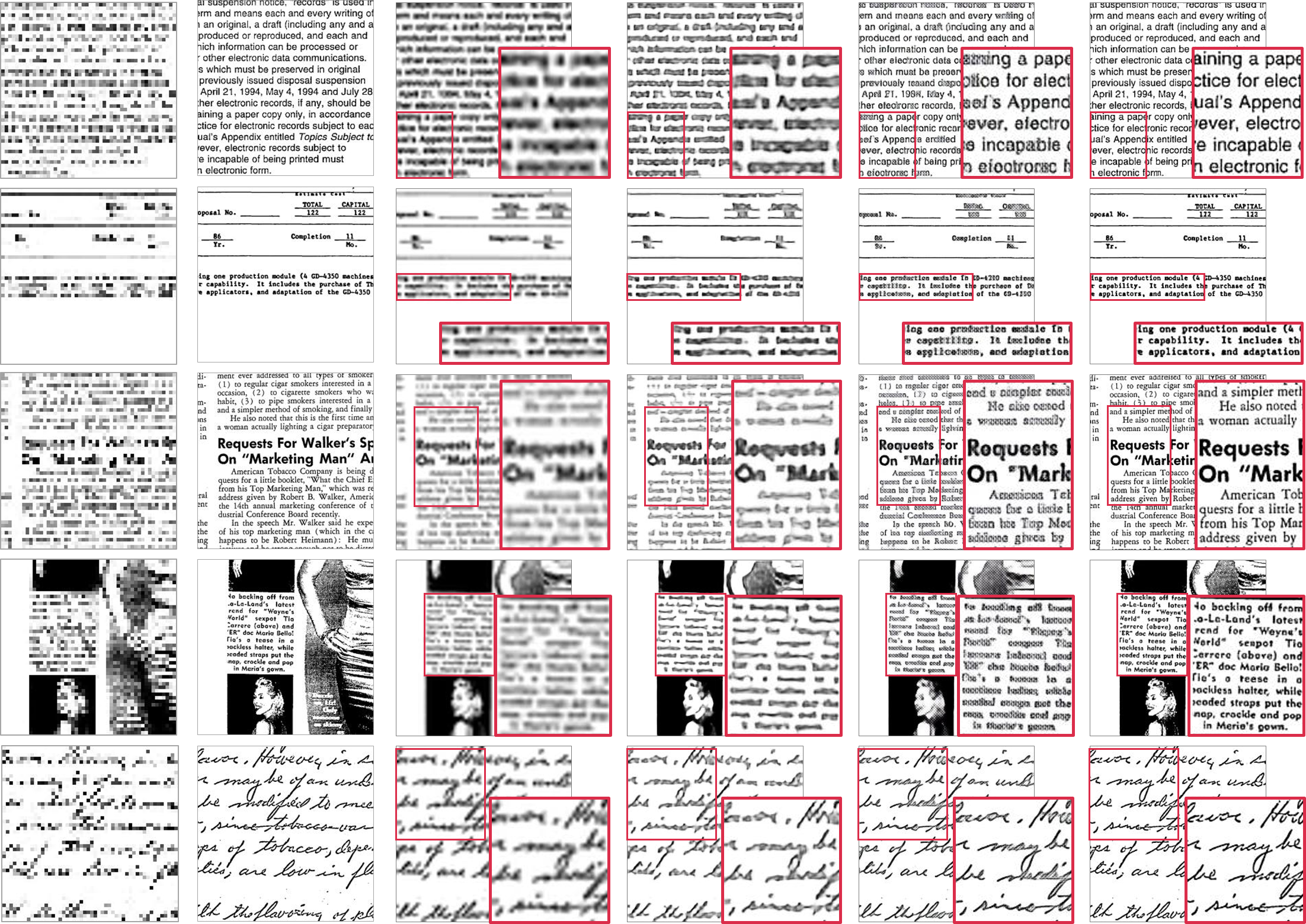}
\caption{
Comparison of $4\times$ results on the RVL-DCIP images. From left to right: low-resolution images (hallucinated in 4$\times$), high-resolution images, results by Bicubic, SRCNN~\cite{Dong2016Image}, SRGAN~\cite{Ledig2016Photo}, and Cascaded DPNets.}
\label{fig:RVL}
\end{figure*}

\begin{table*}[t]
\caption{Overview of the experiment results on both datasets.}
\label{tab:overview}
\centering
{
\begin{tabular}{c||cc||cccc}
\hline
\multirow{2}{*}{Method} & \multicolumn{2}{|c||}{RVL-DCIP Region} & \multicolumn{4}{c}{ICDAR17-Textline}\\
\cline{2-7}
 & PSNR & SSIM& PSNR & SSIM & $S_{LCS}$ & $S_{LD}$\\
\hline\hline
Bicubic & 20.74 & 0.7113& 19.99 & 0.8025 & 0.6109 & 0.5771\\
SRCNN~\cite{Dong2016Image} & 21.98 & 0.7225& 22.77 & 0.8052 & 0.6395 & 0.6080\\
SRGAN~\cite{Ledig2016Photo} & 23.89 & 0.7403& 25.85 & 0.8145 & 0.8617 & 0.8506\\
\hline
Cascaded DPNets & 25.27 & 0.7541& 30.51 & 0.8779 & 0.9134 & 0.9089\\
\hline
\end{tabular}
}
\end{table*}

\begin{figure}[t]
\centering
\includegraphics[width=.9\linewidth]{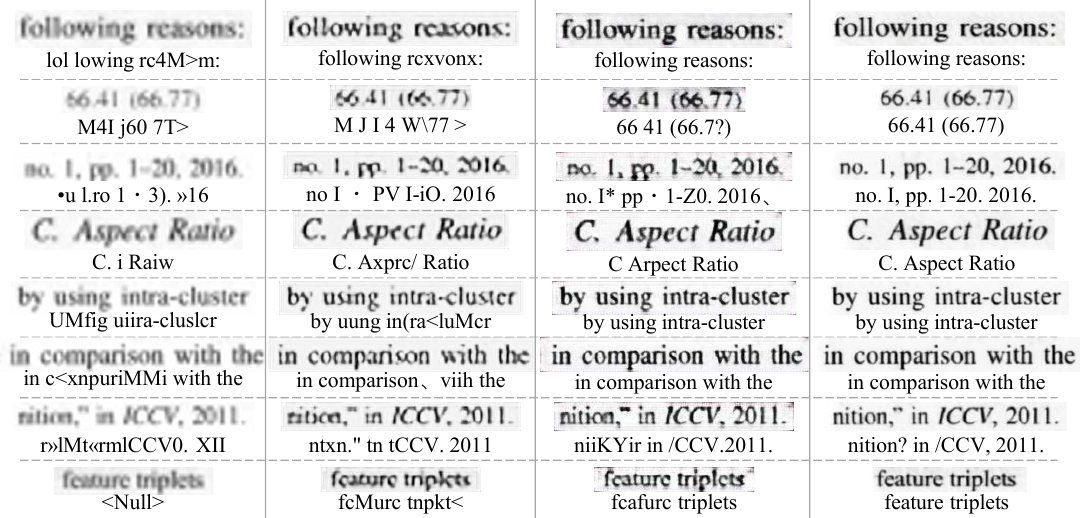}
\caption{Comparison of $4\times$ results on the ICDAR images.
The comparison between testing text patches. Results from left to right: Bicubic, SRCNN, SRGAN, Cascaded DPNets. $\langle$Null$\rangle$ indicates there is no returned result from ABBYY.}
\label{fig:icdar}
\end{figure}

\subsection{Datasets and Evaluation Metric}
\label{subsec:dataset}

To validate the efficiency of the proposed framework, we collect two document image datasets and design two groups of experiments.

\subsubsection{RVL-CDIP Region}

The RVL-CDIP (Ryerson Vision Lab Complex Document Information Processing) dataset~\cite{harley2015icdar} consists of 16 document categories with \emph{25K} document images per category. Among these \emph{400K} grayscale document images, 80\% images are considered as the training set, 10\% images are as the validation set, and the rest are the testing images. In our experiment, we randomly sample \emph{32K} regions with a size of $320\times320$ pixels from the original RVL-CDIP dataset for training, \emph{4K} regions for validation, and \emph{4K} regions as the test set. As the document images are various with different fonts and structures, we focus on both the quantitative and qualitative evaluation of super-resolution results and PSNR as well as SSIM are considered as the metric.

\subsubsection{ICDAR17-Textline}

We also construct a dataset containing textline regions with recognition annotation to evaluate both the super-resolution metrics and the OCR accuracy with super-resolution.
We randomly select 20 pages from the proceeding of ICDAR 2017, print them in the paper, and then scan to full-page digital images with the setting of 300 dpi and 75 dpi. For each page, 30 textline regions are randomly cropped with the text labels by annotators. All of the text region images are divided with 400 patches as the training set and 200 patches as the testing set.

Besides the PSNR and SSIM used in RVL-CDIP experiments, here we also evaluate the OCR performance with the help of image super-resolution.
After the super-resolution process, the output images are sent into a commercial OCR system\footnote{ABBYY Fine Reader 14: \url{https://www.abbyy.com/en-apac/finereader/}}.
Two recognition precision metrics are defined, namely \emph{LCS score} and \emph{Levenshtein score}, with values Fall within the interval
of [0,1]. The LCS score is based on the \emph{Longest Common Subsequence (LCS)}, with the definition as
$$S_{LCS}=\frac{LCS\_length(s,t)}{Maxlen(s,t)},$$
where  $s$ and $t$ indicate the predicted and target text respectively. The LCS score is the ratio of LCS length to the maximum length of the $s$ and $t$, \ie, $Maxlen(s,t)=\max(len(s), len(t))$. It only reaches the maximum value of 1.0 when $s$ is completely the same as $t$.
The Levenshtein score is obtained with the Levenshtein distance. Levenshtein distance, which may also be referred to as edit distance, is a string metric for measuring the difference between two sequences. Therefore, we use the difference between Levenshtein distance and $Maxlen$ to evaluate the similarity between $s$ and $t$, \ie,
$$S_{LD}=1-\frac{Levenshtein\_distance(s,t)}{Maxlen(s,t)}.$$

\subsection{Results and Comparison}

Table \ref{tab:overview} demonstrates the comparison of our full model with state-of-the-art super-resolution approaches.
We compare with classical Bicubic method as well as recent deep learning based models SRCNN~\cite{Dong2016Image} and SRGAN~\cite{Ledig2016Photo}. All of these baseline methods and proposed framework are compared with the same magnification (4$\times$).
Notably, our Cascaded DPNets performs better on both datasets under all the metrics.
Fig. \ref{fig:RVL} and Fig. \ref{fig:icdar} demonstrate qualitative evaluations of our approach on the testing sets.
We succeed in preserving the detail of the text regions in different document types and character fonts, especially when the small characters appear.
However, there are also some failure cases where some characters are extremely small, or fails to identify multiple characters that are adjacent to each other.
The recognition results on the ICDAR17-Textline dataset are also illustrated in Fig. \ref{fig:icdar}. We can observe that combining the proposed Cascaded DPNets with the OCR system can further boost the recognition accuracy.
Generally speaking, the super-resolution results show improvement on the full-reference image quality metrics comparing with baseline methods. Text characters and image details are with high quality for further post-processing such as layout extraction and character recognition.
During inference, the Cascaded DPNet model achieves 75 FPS speed by consuming 2840M memory from an Nvidia GTX Titan Xp GPU with a $128\times128$ LR image as input.

\begin{table}[t]
\caption{Evaluation with different settings on RVL-DCIP.}
\label{tab:edge}
\centering
\begin{tabular}{c|c|c}
\hline
Method & PSNR & SSIM\\
\hline
Cascaded DPNet without Edge & 24.96 & 0.7487\\
Cascaded DPNet with Edge & 25.27 & 0.7541\\
\hline
\end{tabular}
\\
(a) Edge loss\vspace{0.03in}\\
\begin{tabular}{c|c|c}
\hline
Method & PSNR & SSIM\\
\hline
Bicubic ($4\times$) & 20.74 & 0.7113\\
Bicubic ($2\times$) + DPNet ($2\times$) & 21.12 & 0.7218\\
DPNet ($2\times$) + Bicubic ($2\times$) & 22.95 & 0.7361\\
Cascaded DPNet ($4\times$) & 25.27 & 0.7541\\
\hline
\end{tabular}
\\(b) Different cascade structures
\end{table}

\begin{figure}[t]
\centering
\includegraphics[width=\linewidth]{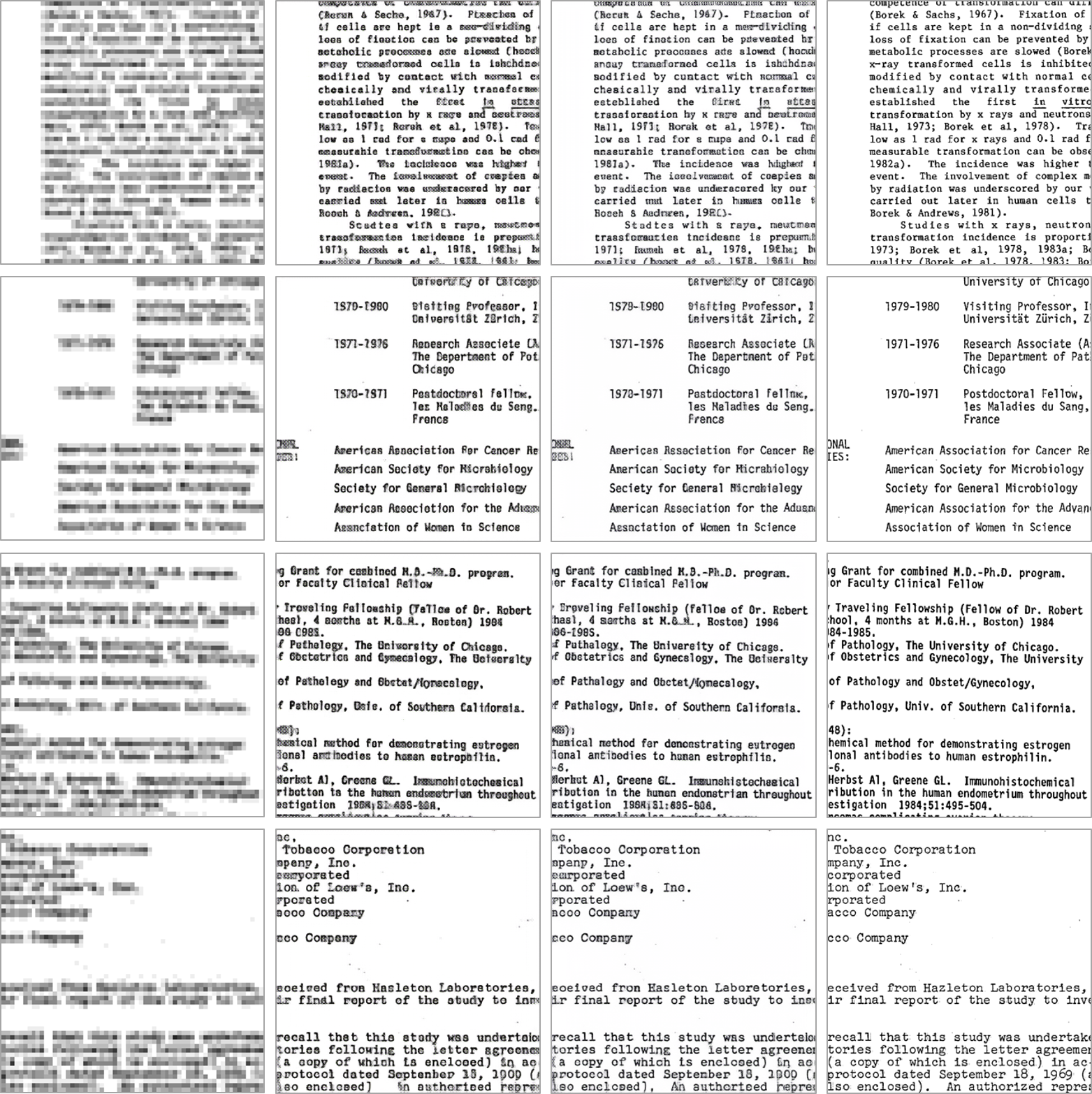}
(a)~~~~~~~~~~~~~~~(b)~~~~~~~~~~~~~~~(c)~~~~~~~~~~~~~~~(d)
\caption{Evaluation of super-resolution results with the edge term. (a) low-resolution images (hallucinated in $4\times$); (b) super-resolved images without edge loss; (c) super-resolved images with edge loss; (d) high-resolution images.}
\label{fig:edge}
\end{figure}

\begin{figure*}[t]
\centering
\includegraphics[width=.8\linewidth]{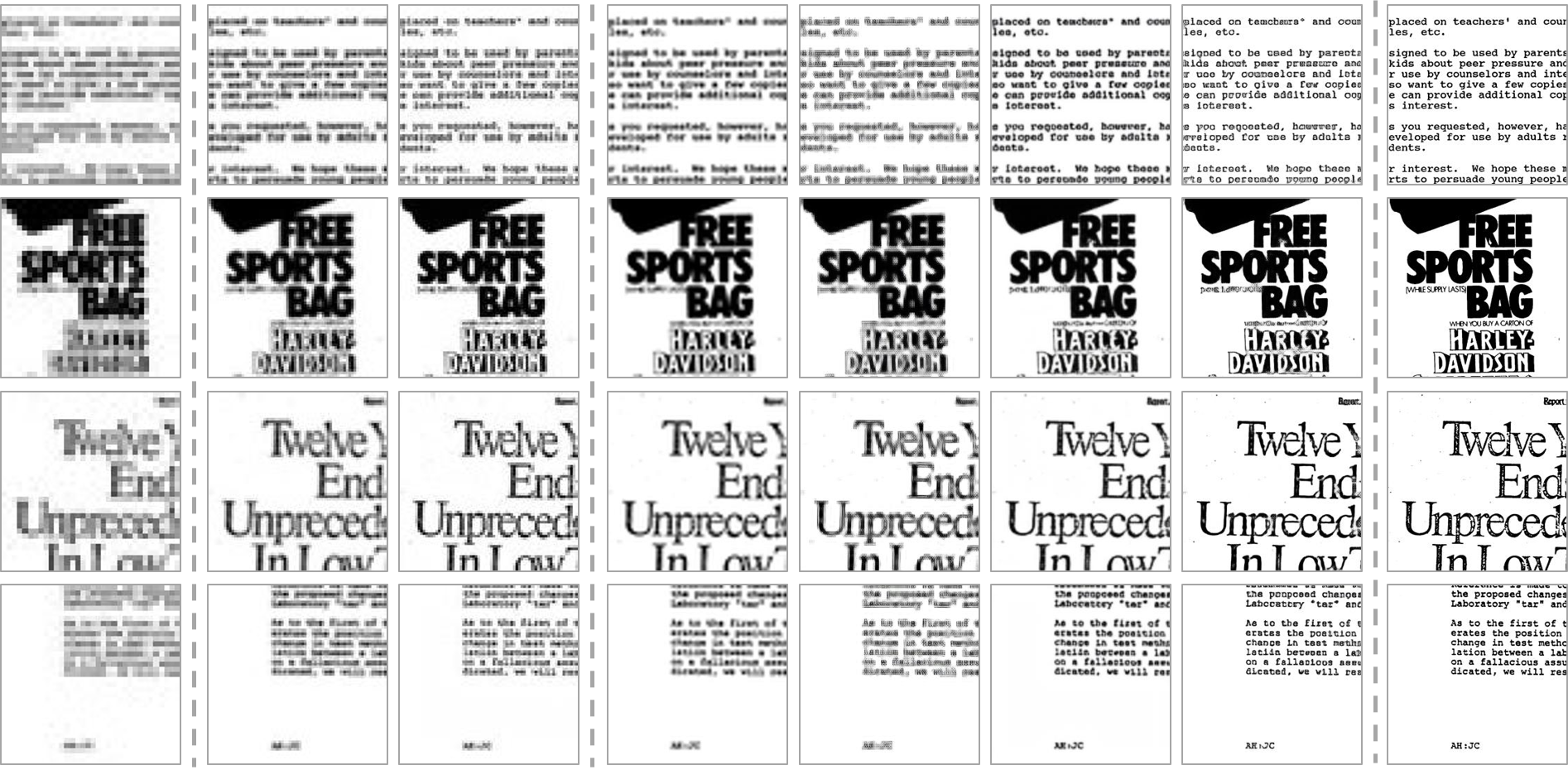}
\caption{Qualitative results of different cascade structures. From left to right: low-resolution images (hallucinated in $4\times$), $2\times$ super-resolution results by bicubic and DPNet (hallucinated in $2\times$), super-resolution results by bicubic, bicubic ($2\times$) + DPNet ($2\times$), DPNet ($2\times$) + bicubic ($2\times$), and Cascaded DPNet ($4\times$), and high-resolution images.}
\label{fig:bicubic}
\end{figure*}

\subsection{Ablation Study}

In this subsection, we evaluate the alternative implementations for the document image super-resolution. We report results on the RVL-CDIP Region dataset as it is larger and more diversified than ICDAR17-Textline.

Recall that the edge term is computed to represent the edge information, which is of great importance as mentioned in Section \ref{fig:loss}. The super-resolved images and their corresponding metrics with or without the edge loss are shown in Fig. \ref{fig:edge} and Table \ref{tab:edge}(a).
The cascaded networks without edge loss outperform the SRGAN framework, indicating the effectiveness of cascade architecture on document images.
We observe performance gains when adding the edge term, and the super-resolved text regions are with better contour and more clear characters that are helpful for further recognition.

We also evaluate the effect of components within the cascade super-resolution structure. Fig. \ref{fig:bicubic} and Table \ref{tab:edge}(b) demonstrate the comparison with replacing the DPNet with bicubic. Quantitatively speaking, the model with DPNets performs the best among different cascade settings.
The multiple stages of DPNet introduce a 10.5\% gain on PSNR over the cascade of Bicubic and DPNet, and a significant improvement of the SR results as illustrated in Fig. \ref{fig:bicubic}.

\section{Conclusions}
\label{sec:conclusion}

We have introduced Cascaded DPNets, a deep super-resolution framework for the document images. Detail-Preserving Network with small magnification is able to preserve the content and enhance the edge of the characters. The cascade of the networks is assembled into a pipeline model with a larger magnification. Through an extensive set of document super-resolution experiments, we have shown that Cascaded DPNets is more effective than the baseline deep learning approaches, generating very competitive results from the low-resolution document images.

\bibliographystyle{IEEEtran}
\bibliography{total}

\end{document}